\documentclass[
twocolumn
]{ceurart}

\usepackage{listings}
\usepackage{graphicx}
\usepackage{natbib}
\usepackage{doi}
\usepackage[dvipsnames]{xcolor}
\usepackage{todonotes}
\usepackage{caption}
\usepackage{subcaption}
\usepackage{url}
\begin{document}

\copyrightyear{2022}
\copyrightclause{Copyright for this paper by its authors.
  Use permitted under Creative Commons License Attribution 4.0
  International (CC BY 4.0).}

\conference{SwissText}

\title{Text-to-Speech Pipeline for Swiss German - A comparison}

\author[1]{Tobias Bollinger}[%
email=tobias.bollinger@students.fhnw.ch
]
\cormark[1]
\address[1]{University of Applied Sciences and Arts Northwestern Switzerland, Windisch}

\author[2]{Jan Deriu}[%
email=jan.deriu@zhaw.ch
]
\address[2]{Zurich University of Applied Sciences, Winterthur}

\author[1]{Manfred Vogel}[%
email=manfred.vogel@fhnw.ch,
]

\cortext[1]{Corresponding author.}

\begin{abstract}
In this work, we studied the synthesis of Swiss German speech using different Text-to-Speech (TTS) models.
We evaluated the TTS models on three corpora, and we found, that VITS models performed best, hence, using them for further testing.
We also introduce a new method to evaluate TTS models by letting the discriminator of a trained vocoder GAN model predict whether a given waveform is human or synthesized.
In summary, our best model delivers speech synthesis for different Swiss German dialects with previously unachieved quality.
\end{abstract}

\begin{keywords}
  Speech Synthesis \sep
  Text to Speech \sep
  Swiss German \sep
  low resource
\end{keywords}

\maketitle

\section{Introduction}
Text-to-Speech (TTS) describes the task of synthesizing raw waveform of speech for a given text. While the field is making staggering progress for many large-resource languages, the work on Swiss German is still sparse. There are several reasons for this. First, Switzerland is a small country and around 60\% speak Swiss German~\footnote{\url{https://www.eda.admin.ch/aboutswitzerland/de/home/gesellschaft/sprachen/die-sprachen---fakten-und-zahlen.html}} making it a low resource setting. Second, there is no standardized text form, thus, the system has to handle standard German input to create Swiss German audio. Third, there is no single Swiss German language, on the contrary, there are many  dialects that differ quite significantly. The dialects can also differ significantly in their grammar, vocabulary and phonetics. Thus, the setting for TTS for Swiss German is challenging. 

On the other hand, there was a push in recent years to collect high-quality data for Swiss German \cite{SwissDial,SDS-200,SlowSoftGmbH2019Nov}. This opens up for the possibility of training higher quality TTS systems, and also textural translation systems from standard German to different Swiss German dialects. For instance, \cite{SDS-200} collected around 200 hours of Swiss German speech with the corresponding standard German translation, and \cite{SwissDial} collected around 30 hours of high-quality Swiss German audio and provide Swiss German texts to specifically train TTS systems. 

In this work, we make two contributions. First, we assess the quality of the variational inference with adversarial learning for end-to-end Text-to-Speech (VITS)~\cite{vits} when trained on three different data sets, namely \cite{SwissDial,SDS-200,SlowSoftGmbH2019Nov}. Second, we train a textual translation system for standard German to Swiss German text (although there is no formal grammar for it), which is used as prepossessing to input Swiss German text to the TTS system. 

Our results show that using high-quality data and using a translation system for translating standard German text to Swiss German text, yields high quality synthesized speech. That is, it achieves an average of 4.1 on a 5-point scale (MOS), which is considered high quality. 



\section{Related Work}

\subsection{Speech synthesis in Swiss German}
Text to Speech models are prominent for many languages like English, Japanese, Korean or German, but for Swiss German less work is done. In 2016, SlowSoft GmbH developed a speech synthesis system for Grisons and Romansh \citep{SlowSoftGmbH2019Nov}. The online service Narakeet, a text-to-speech video maker, also supports one Swiss German speaker \citep{Narakeet2022Dec}. \citep{SwissDial} published a Swiss German Corpus and also trained a Tacotron2 \citep{tacotron2} model for Swiss German speech synthesis.

\subsection{Deep learning-based synthesis}
Current TTS methods can be categorized into cascaded and end-to-end methods. In cascaded TTS systems like Tacotron2 \citep{tacotron2} or Fastspeech2 \citep{Fastspeech2}, the task of synthesis is divided into two steps. First, generating an intermediate representation of Mel spectrograms out of the text with a so-called acoustic model. Second, using the Mel spectrograms to generate the final wave form with a vocoder. End-to-end TTS systems try to jointly optimize the acoustic model and the vocoder. This system is used by the VITS model \citep{vits}. To get more control over the produced synthesis, zero-shot models like YourTTS \citep{YourTTS} or Vall-E \citep{VALL-E} provide a method to produce speech out of text and a short speech input to imitate the voice of the speaker of the provided speech input.

\section{Training Data} \label{sec:corpora}
\begin{table}[!t]
	\caption{Data set size used for training: SDS-200, SwissDial, SlowSoft.}
	\centering
	\begin{tabular}{lll}
		SDS-200\\
		\toprule
		Set & Fraction & Samples \\
		\midrule
		Train & 0.9 & 12'948 \\
		Valid & 0.05 & 720 \\
		Test & 0.05 & 720 \\
		\bottomrule
	\end{tabular}
	\quad
	\begin{tabular}{lll}
		SwissDial \\
		\toprule
		Set & Fraction & Samples \\
		\midrule
		Train & 0.9 & 27'828 \\
		Valid & 0.05 & 1'546 \\
		Test & 0.05 & 1'547 \\
		\bottomrule
	\end{tabular}
	\quad
	\begin{tabular}{lll}
		Slowsoft \\
		\toprule
		Set & Fraction & Samples \\
		\midrule
		Train & 0.9 & 2'061  \\
		Valid & 0.05 & 115 \\
		Test & 0.05 & 115  \\
		\bottomrule
	\end{tabular}
	\label{tab:datasetSize}
\end{table}

The focus of this work is to benchmark how well the end-to-end training of VITS performs on Swiss German data sets. For this, we used three different corpora, which differ on the quality of the audio and the number of samples per speaker.  


\subsection{SDS-200}
The SDS-200 corpus contains 200 hours of Swiss German audio recordings from all dialects, along with their Standard German transcriptions. Note that in contrast to classical TTS corpora, the SDS-200 contains a large variety of different speakers as the primary purpose of the corpus is training Speech-to-Text models.
The recordings represent one sentence each and have been crowd-sourced from the Swiss population during the "Schweizer Dialektsammlung" \citep{SDS-200}. The audio files are compressed and there is a lot of background noise in the audios such as mouse clicking at the end of the audio. However, this is the largest corpus in our set, and it showed to be useful for training speech-to-text systems. 

\subsection{SwissDial}
SwissDial is an annotated corpus of spoken Swiss German across 8 major dialects (AG, BE, BS, GR, LU, SG, VS, ZH). The data set includes around 3 hours of high quality audio per dialect, together with Swiss German and Standard German transcripts \citep{SwissDial}. The audio files have been recorded in a quiet room in front of a high-quality recording set-up to obtain clean recordings. We use the Swiss German and standard German transcripts to train the textual translation model. Note that in SwissDial the Grisons dialect has an extra 6 horus of audio samples compared to the other dialects. 

\subsection{Slowsoft}
SlowSoft builds speech technology components with a focus on minority languages and dialects. The data, kindly provided for our research, contains phonetic transcriptions created by professionals, in the Grisons (GR) dialect \citep{SlowSoftGmbH2019Jun}.

Table~\ref{tab:datasetSize} contains the overview of the data set sizes, which we used to train the TTS pipeline. 


\section{Pipeline}
The pipeline consists of two main parts: first, a T5~\cite{reffel2020t5} model fine-tuned on translating standard German text to Swiss German text, and second the VITS model trained on the data sets mentioned in the previous section. 

\subsection{T5 - German to Swiss German}
The standard German to Swiss German text translation is based on a T5 model~\cite{reffel2020t5} fine-tuned on the SwissDial data set~\citep{SwissDial}. As this data set contains for each sample the German sentence alongside the Swiss German sentences for eight different dialects, we train the translation model to translate to the various dialects. For this, we used the Huggingface~\cite{wolf-etal-2020-transformers} implementation. To allow the model to handle the different dialects, we prepended the dialect tag to the input.

\subsection{VITS - Swiss German Text to Speech}
The Conditional Variational Autoencoder with Adversarial Learning for End-to-End Text-to-Speech is a parallel end-to-end TTS method. It adapts variational inference with normalizing flows and an adversarial training process \citep{vits}. For the experiment, the ESPNet Framework was used \citep{watanabe2018espnet}, which implements the VITS architecture.

The input text was split into a sequence by character or word. When a split by characters was used, it generalized way better than when the text was split by words and therefore produced better results.

We also tested a Grapheme-to-Phoneme (G2P) conversion for German text which uses eSpeak as backend \citep{eSpeak}. But because the rule-based G2P algorithm only works for German and not Swiss German, it results in worse synthesized speech.

Different preprocessing steps were tested, for SDS-200 a noise reduction was applied and for all corpora silence at the start and end were trimmed. The reduction of noise resulted in a more unnatural result, therefore only the trim off the silence was used for preprocessing.

To allow the synthesis of different dialects, a speaker embedding with x-vectors \citep{xvector} with an embedding dimension of 512 has been used.

Thus, we train three VITS models, one for each of the three different corpora described in \ref{sec:corpora}.


\section{Evaluation Metrics}
The evaluation of Text-to-Speech is a challenging topic. Therefore, we perform an extensive automated evaluation using a variety of different metrics, as well as a human evaluation using a 5-point Mean Option Score (MOS). For the T5 model, we use the BLEU score~\cite{papineni2002bleu} to evaluate and propose a qualitative analysis. 

\paragraph{MCD}
Mel-Cepstral Distortion~\citep{MCD} is a measure which calculates how different two sequences of Mel cepstral are. The idea is, the smaller the MCD is between synthesized and natural Mel cepstral sequences, the closer the synthesized speech is to the natural speech. This is by no means a perfect measurement, but it can be a useful indicator, especially combined with other metrics. 

\paragraph{log-F0 RMSE}
Log-F0 Root-Mean-Square Error is a measure which computes the root-mean-square error of the log-F0 of the signal. Where F0 is the fundamental frequency which is closely related to the pitch.

\hfill \break
For both, the MCD and log-F0 RMSE dynamic time-warping (DTW) is applied to match the different length of the ground-truth and the generated signal.

\paragraph{CER and WER}
The Character Error Rate (CER) and the Word Error Rate (WER) are calculating the modifications needed to transform the ground-truth text to the generated text. A modification is defined as either a substitution ($S$), a deletion ($D$) or an insertion ($I$). When there is no modification needed, we call it a Word/Character Hit ($H$) \citep{Morris2004Oct}. For CER the calculation is done on a character basis:

\begin{equation}
	CER = \frac{S + D + I}{H + S + D}
\end{equation}
$H + S + D$ is the number of characters in the ground-truth text.

For WER the calculation is done on a word basis, the same equation can be used, but the modifications are counted per word and not per character. 

\begin{equation}
	WER = \frac{S + D + I}{H + S + D}
\end{equation}

Before calculating the error rate, both texts are preprocessed:
\begin{itemize}
	\item Convert the text to lower case
	\item Remove multiple spaces
	\item Trim white spaces at the start and end
	\item Remove the three characters: \textit{«}, \textit{»} and \textit{,}
	\item Remove punctuation
\end{itemize}

\paragraph{BLEU}
The Bilingual Evaluation Understudy score counts the matching n-grams in the ground-truth text to the generated text. A perfect match results in a score of 1.0, whereas a perfect mismatch results in a score of 0.0.

\hfill \break
MCD and log-F0 RMSE reflect speaker, prosody, and phonetic content similarities, whereas CER, WER and BLEU the intelligibility.

\paragraph{Vocoder Discriminator Score}
A further method -- developed in this work -- to measure the quality of the speech synthesis is to use a trained vocoder which uses Generative Adversarial Networks (GAN). The discriminator tries to discriminate fake from real waveforms. The intuition is, the better our synthesized speech can fool the discriminator, the better the model performed. To evaluate this, we can take the discriminator or discriminators of the trained model and let them predict if the speech input is fake or real. The discriminators will output a value for each waveform step between $0$ and $1$ where a bigger number means the discriminator is more confident that the waveform step is real. For getting the score, the discriminator outputs are concatenated to $\hat{y}$, then the mean squared error between $\hat{y}$ and a zero vector with the same dimension is calculated. The resulting $loss$ will be transformed to a $score$ by subtracting the $loss$ from 1.

\begin{equation}
	\begin{split}
	loss = & \frac{1}{n} \sum_{i=1}^{n} (\hat{y}_i - 1)^2 \\
	score = & 1 - loss
	\end{split}
\end{equation}

Two models were trained for the evaluation, a Parallel WaveGAN \citep{ParallelWaveGan} model and a MelGAN \citep{MelGAN} model; where the Parallel WaveGAN has one discriminator and the MelGAN has three. For the MelGAN we calculate the score for each discriminator and then we take the mean of the three scores. Booth models were trained on the SDS-200 and on the SwissDial corpus.

\begin{figure}[h]
	\centering
	\begin{subfigure}{0.47\textwidth}
		\includegraphics[width=\textwidth]{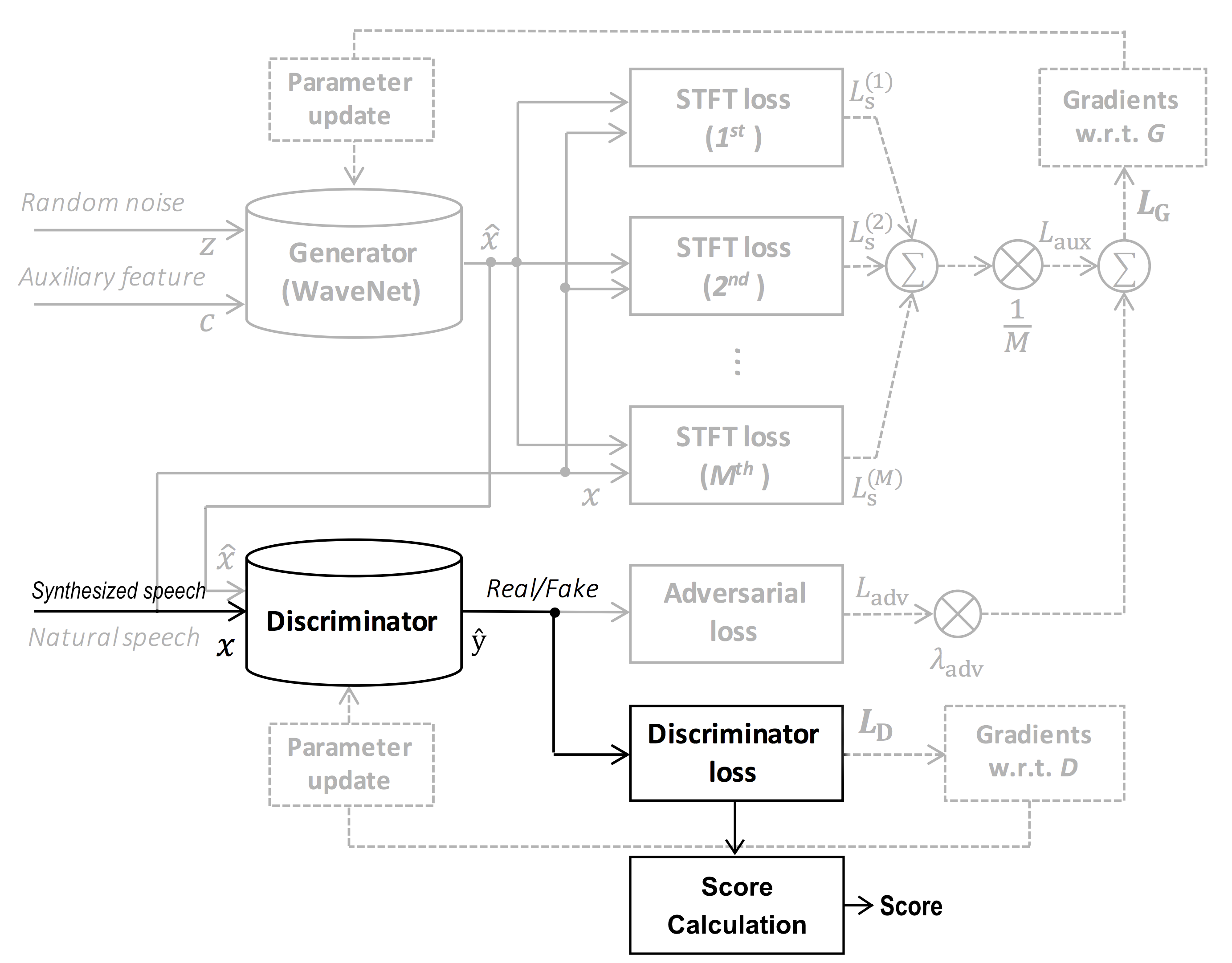}
		\caption{Parallel WaveGAN score calculation. Base on the illustration form \citep[S.2]{ParallelWaveGan}.}
		\label{fig:pganScore}
	\end{subfigure}%
	\hspace{0.05\textwidth}
	\begin{subfigure}{0.47\textwidth}
		\centering
		\includegraphics[width=\textwidth]{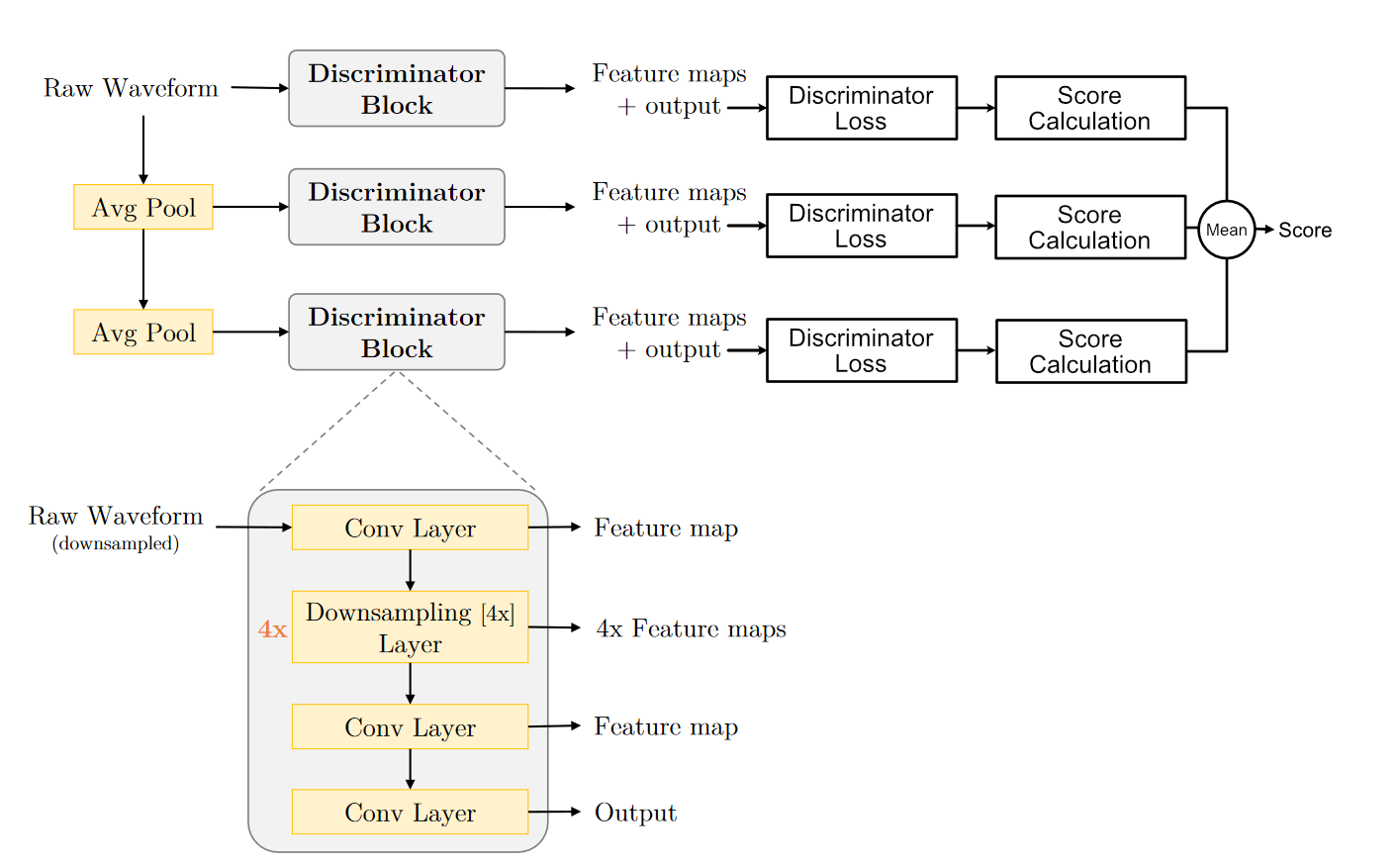}
		\vspace{1cm}
		\caption{MelGAN score calculation. Base on the illustration form \citep[S.4]{MelGAN}.}
		\label{fig:melganScore}
	\end{subfigure}%
\end{figure} 

\begin{table}[t!]
	\caption{Evaluation metrics from left to right: SDS-200, SwissDial, SlowSoft. The arrow at the start symbolizes if smaller values are better $\downarrow$ or if greater values are better $\uparrow$. {\color{ForestGreen}Green} represents the best result.}
	\centering
	\begin{tabular}{ll}
		SDS-200 \\
		\toprule
		Metric Name     & Error / Score \\
		\midrule
		$\downarrow$ MCD& \color{ForestGreen} 7.50 ± 1.63 \\
		$\downarrow$ log-F0 RMSE& \color{ForestGreen}0.22 ± 0.09 \\
		$\uparrow$ PGAN Score& 0.73 \\
		$\uparrow$ MelGAN Score& 0.35\\
		$\downarrow$ CER& 0.27 \\
		$\downarrow$ WER& 0.43 \\
		$\uparrow$ BLEU& 0.39 \\
		\bottomrule
	\end{tabular}
	\quad
	\begin{tabular}{ll}
		SwissDial \\
		\toprule
		Metric Name     & Error / Score \\
		\midrule
		$\downarrow$ MCD &  9.29 ± 1.11 \\
		$\downarrow$ log-F0 RMSE & 0.31 ± 0.11 \\
		$\uparrow$ PGAN Score& \color{ForestGreen} 0.75 \\
		$\uparrow$ MelGAN Score& \color{ForestGreen} 0.69 \\
		$\downarrow$ CER & \color{ForestGreen} 0.15 \\
		$\downarrow$ WER & \color{ForestGreen} 0.28 \\
		$\uparrow$ BLEU & \color{ForestGreen} 0.54 \\
		\bottomrule
	\end{tabular}
	\quad
	\begin{tabular}{ll}
		Slowsoft \\
		\toprule
		Metric Name     & Error / Score \\
		\midrule
		$\downarrow$ MCD & 8.27 ± 0.66 \\
		$\downarrow$ log-F0 RMSE & 0.25 ± 0.08 \\
		$\uparrow$ PGAN Score&  0.74 \\
		$\uparrow$ MelGAN Score& 0.57 \\
		\bottomrule
	\end{tabular}
	\label{tab:eval}
\end{table}

\begin{table*}[!t]
\centering
\small
\resizebox{.9\textwidth}{!}{
\begin{tabular}{l|l}
Dialect  & Utterance       \\
\hline
\textsc{DE}    & In Gegenrichtung wurde jedoch eine durchgehende Leitplanke installiert. \\
\hline
\textsc{ZH}   & In Gägerichtig isch aber e durchgehendi Litplank installiert worde.      \\
\textsc{SG}   & In Gegenrichtig isch aber e durchgehendi Leitplank installiert worde.    \\
\textsc{BE}   & In Gägärichtig isch aber ä durchgehendi Litplanki inschtalliert wordä.   \\
\textsc{GR}   & In Gegarichtig isch jedoch a durchgegehndi Leitplanka installiart worda. \\
\textsc{VS}   & In Gägurichtig isch aber en durchgehendi Leitplanka installiert wordu.  \\
\textsc{BS}   & In Gegerichtig isch aber ä durchgenedi Leitplank installiert worde. \\ 
\textsc{AG}   & E Gägerechtig esch aber en dorchgehendi Leifantworte installiert worde. \\ 
\textsc{LU}   & En Gägesiitig esch aber en döregehendi Liitplank installiert worde. \\ 
\hline
\textsc{DE}    & Zunächst kommen aber ohnehin die Rezepte des Bundesrates ins Parlament. \\
\hline
\textsc{ZH}   & Zerst chömed aber sowieso d Rezäpt vom Bundesrat is Parlamänt.   \\
\textsc{SG}   & Zersch chömmed aber trotzdem dRefekt vom Bundesrot is Parlament.    \\
\textsc{BE}   & Zersch chömä aber trotzdäm d Projekt vom Bundesrat is Parlamänt.   \\
\textsc{GR}   & Zerst kömmen aber sowieso d Rezept vum Bundesrot ins Parlament.	 \\
\textsc{VS}   & Zersch chämunt aber sowieso d'Reschtoff vam Bundesrat ins Parlamänt.  \\
\textsc{BS}   & Zerst kömme aber sowieso d Rezäkt vom Bundesroot ins Parlament. \\ 
\textsc{AG}   & Zerscht chömme aber sowieso d'Reschpondänt fom Bondesrot es Parlamänt. \\ 
\textsc{LU}   & Zerscht chömed aber sowieso d Rezäpt vom Bondesrot is Parlamänt. \\ 
\hline
\textsc{DE}    & Er führte auch das Servieren von Eiscreme bei nachmittäglichen Empfängen ein. \\
\hline
\textsc{ZH}   & Er hät au s Serviere vo Glacé bi nahmittägliche Empfäng igfüert.  \\
\textsc{SG}   & Er het au sServiere vo Iiscreme bi nochemtägliche Empfäng iigführt.   \\
\textsc{BE}   & Är het o s Servierä vo Glacä bi nachmittäglechä Empfäng iigfüehrt.	   \\
\textsc{GR}   & Er het au ds Serviara vu Iiscreme bi nochmittäglicha Empfäng igfüahrt. \\
\textsc{VS}   & Är het öi z'Servieru va Iischcreme bi nahmittaglichu Empfängi ihgfiehrt. \\
\textsc{BS}   & Är het au s Serviere vo Glace bi nochmittägliche Empfäng ihgfüehrt. \\ 
\textsc{AG}   & Er het au s'Serviere fo Iscreme be nachhertägliche Emfäng ihgfüehrt. \\ 
\textsc{LU}   & Er hed au s Serviere vo Wiisscreme be nochmettägliche Emfäng igfüehrt. \\ 
\hline
\end{tabular}
}
\caption{Three randomly picked examples of standard German text translated to each of the dialects in the Siwss Dial corpus.}
\label{tab:selected}
\end{table*}

\section{Results}
In this section we present the results of the pipeline. For this we show the scores of the T5 model and the different scores of the VITS models trained on the three data sets. 
\subsection{T5 - Results}
The T5 model achieves a BLEU score of 50.47 on the test split of the SwissDial corpus over all the eight dialects, which is very promising. However, the BLEU score is not fully representative. Thus, we add a qualitative analysis. Table~\ref{tab:selected} shows three examples of the Swiss German texts generated from the standard German input. We note that the past tense is handled correctly in these cases. Since Swiss German does not use the simple past, all the past tenses need to be rewritten to past progressive tense. In the example, "In Gegenrichtung wurde jedoch eine durchgehende Leitplanke installiert." is translated correctly to "In Gegarichtig isch jedoch a durchgegehndi Leitplanka installiart worda." Similarly the last example changes "führte" to "hät ... igfüert". Most discrepancies between the ground truth and the generated Swiss German sentences stem from the fact that different spellings were used. For instance, in the first example, the ZH groundtruth would have expected to start with "Ide Gägerichtig..." instead of "In Gägerichtig...". However, the generated utterance is not wrong. The evaluation is challenging since Swiss German has no writing rules and the specific spelling is, thus, very individual. There are also real mistakes, such as in the last example the LU translation uses the word "Wiisscreme", which is clearly wrong. 
 \subsection{VITS - Results}
For evaluating CER, WER and BLEU an internally provided ASR fairseq model was used. The model reached on SDS-200 with GT speech a BLEU of 0.698, a WER of 0.18 and a CER of 0.12. Because the Slowsoft model uses phonemes, it is not possible to evaluate the CER, WER and BLEU metrics.

The SDS-200 model performed best on the MCD and log-F0 RMSE metrics. The corpus SDS-200 contains some noise which is an easy task for the model to reproduce correctly. We think this is also the reason the model performed the best on the MCD and log-F0 RMSE metrics which compares the ground-truth signal with the synthesized. But on all other metrics, the SwissDial model performed best. The Slowsoft model has significantly less training data than the other two corpora, but does not show a significant drop in performance. It even achieves better results on MCD and log-F0 RMSE compared to the SwissDial model.

\subsection{Human Evaluation - MOS}
\begin{figure}[t!]
	\centering
	\begin{subfigure}{0.5\textwidth}
		\includegraphics[width=\textwidth]{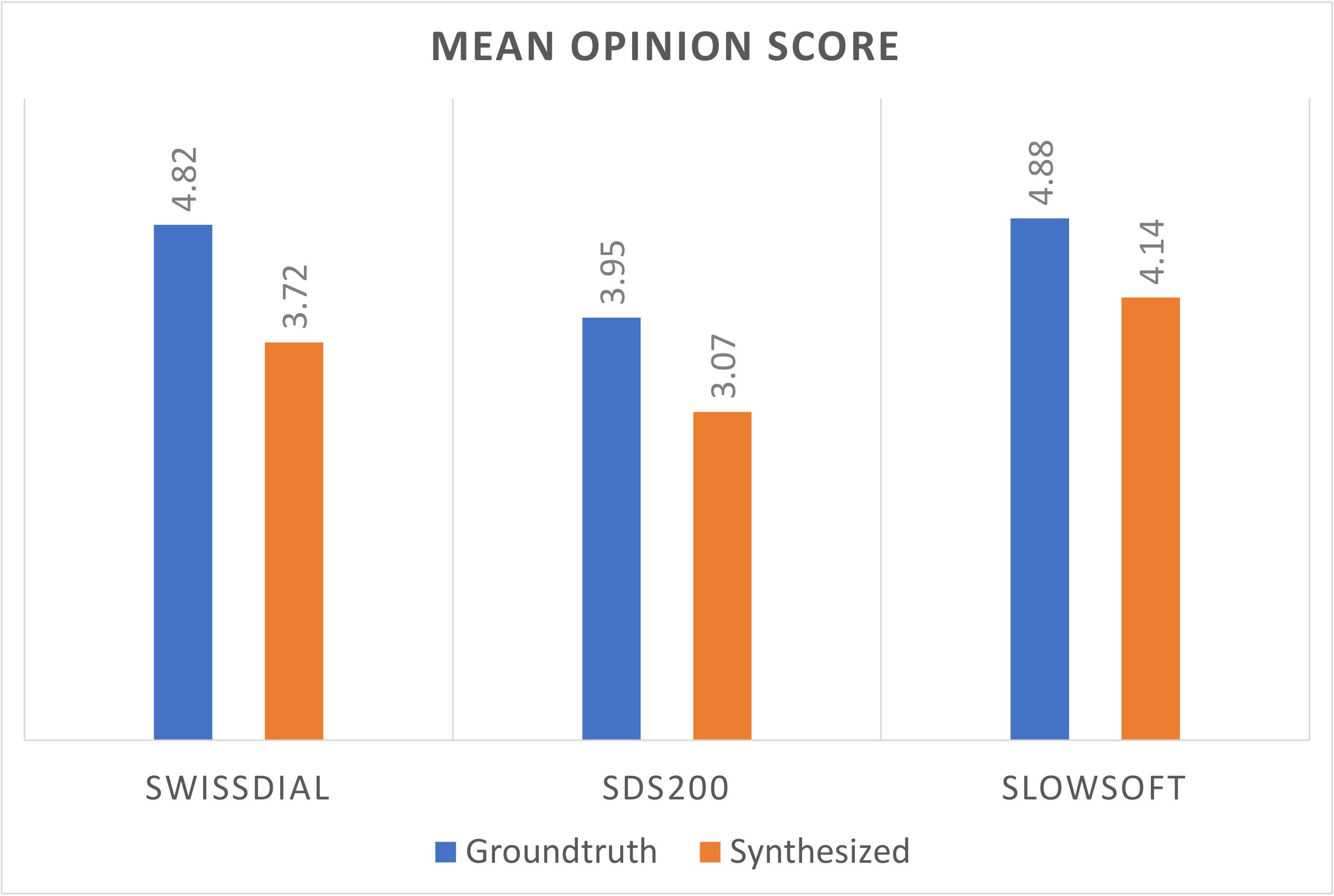}
	\end{subfigure}%
 \caption{MOS Results of the TTS human evaluation.}
  \label{fig:mos}
\end{figure} 
Since automated evaluation is flawed, we also run a human evaluation. For this, we randomly select 15 samples, which we synthesize into waveforms using the three different VITS models. We then let human rate the audio on a scale from 1 to 5 with the following classification:
\begin{itemize}
    \item[5]: Excellent: Neural speech synthesis for Swiss German.
    \item[4]: Good: Through attentive listening, speech can be perceived without effort.
    \item[3]: Fair: Speech can be perceived with slight effort.
    \item[2]: Poor: It takes great concentration and effort to understand the speech.
    \item[1]: Bad: Despite great effort, you cannot understand the speech.
\end{itemize}

We let each sample be rated by 7 native speakers. Alongside the synthesized audio, we also let the groundtruth audio be rated as a reference. In Figure~\ref{fig:mos} the results of the human evaluation are shown. We see that the groundtruth audio are still significantly higher rated than the synthesized audios. Consistent with our observations, the groundtruth SDS-200 audio samples were rated worse compared to the high-quality SwissDial and Slowsoft samples. The synthesized samples show the same trends. The VITS model trained with the high-quality Slowsoft samples performs best with an average MOS score of 4.1, which is higher than the groundtruth samples of the SDS-200 corpus. Analogous to above, the VITS model trained on SDS-200 samples performs worst, with a score of 3.1. 

In summary, the results show that TTS already works very well for Swiss German, provided that the training material is of high quality. Thus, it is best to use a small amount of data, which is of highest quality, than using a large dataset of low-quality samples. Furthermore, the human evaluation shows that the automated metrics are not indicative of the final quality of the generated audio.

\section{Conclusion}
In this work, we compared three VITS models trained on different data sets to synthesize Swiss German speech from standard German texts. For this, we build a two-step pipeline where we first use a fine-tuned T5 model to translate the standard German text to Swiss German text, and then the VITS model that synthesizes Swiss German audio from the text. We evaluated the quality of the audio with automated metrics and a human evaluation. The results showed that the VITS model trained on the highest quality data performs best. Furthermore, the quality is good enough to be considered high quality audio samples, even surpassing the groundtruth samples of another corpus. This shows that TTS methods are applicable to the case of Swiss German provided that the training data is of high quality and a text translation from standard German to Swiss German text is available.

\begin{acknowledgments}
This work was supported by Swiss National Science Foundation within the project "End-to-End Low-Resource Speech Translation for Swiss German Dialects (E2E\_SG)" [205121\_200729/1].
\end{acknowledgments}

\bibliography{sample-ceur}

\appendix

\end{document}